\documentclass[letterpaper]{article} 
\usepackage{aaai24}  
\usepackage{times}  
\usepackage{helvet}  
\usepackage{courier}  
\usepackage[hyphens]{url}  
\usepackage{graphicx} 
\usepackage{natbib}  
\usepackage{caption} 
\usepackage{algorithm}
\usepackage{algorithmic}
\usepackage{xcolor}
\usepackage{soul}
\definecolor{frenchblue}{rgb}{0.0, 0.45, 0.73}

\usepackage{newfloat}
\usepackage{listings}
\DeclareCaptionStyle{ruled}{labelfont=normalfont,labelsep=colon,strut=off} 
\floatstyle{ruled}
\newfloat{listing}{tb}{lst}{}
\floatname{listing}{Listing}
\usepackage{booktabs}
\usepackage{multirow}
\usepackage{multicol}
\usepackage{bbding}
\usepackage{amssymb}
\usepackage{amsmath}
\usepackage{float}
\usepackage{afterpage}
\usepackage{color}
\definecolor{SeaGreen}{RGB}{46,139,87}
\definecolor{IndianRed}{RGB}{240,99,99}

\title{ViTree: Single-Path Neural Tree for Step-Wise \\
        Interpretable Fine-Grained Visual Categorization}
\author {
    Danning Lao,
    Qi Liu†,
    Jiazi Bu†,
    Junchi Yan,
    Wei Shen*
}
\affiliations {
    MoE Key Lab of Artificial Intelligence, AI Institute, Shanghai Jiao Tong University\\
    \{laodanning, purewhite, bujiazi001, yanjunchi, wei.shen\}@sjtu.edu.cn\\
}

\usepackage{bibentry}

\newcommand{\mymodell}{ViTree }
\newcommand{\mymodel}{ViTree}
\begin{document}
\maketitle

\begin{abstract}
    As computer vision continues to advance and finds widespread applications across various domains, the need for interpretability in deep learning models becomes paramount. Existing methods often resort to post-hoc techniques or prototypes to explain the decision-making process, which can be indirect and lack intrinsic illustration. In this research, we introduce \mymodel, a novel approach for fine-grained visual categorization that combines the popular vision transformer as a feature extraction backbone with neural decision trees. By traversing the tree paths, \mymodell effectively selects patches from transformer-processed features to highlight informative local regions, thereby refining representations in a step-wise manner. Unlike previous tree-based models that rely on soft distributions or ensembles of paths, \mymodell selects a single tree path, offering a clearer and simpler decision-making process. This patch and path selectivity enhances model interpretability of \mymodel, enabling better insights into the model's inner workings. Remarkably, extensive experimentation validates that this streamlined approach surpasses various strong competitors and achieves state-of-the-art performance while maintaining exceptional interpretability which is proved by multi-perspective methods. Code can be found at \url{https://github.com/SJTU-DeepVisionLab/ViTree}.
\end{abstract}

\section{Introduction}



The vision transformer~\cite{Dosovitskiy2020AnII} is a significant advancement in computer vision, employing self-attention mechanisms to excel in various visual tasks~\cite{Yu2021MetaFormerIA, Chou2023FinegrainedVC}. However, their lack of interpretability hinders fairness, transparency, and accountability in AI systems~\cite{Akhtar2023ASO}. In computer vision, post-hoc techniques interpret models after training to comprehend their decisions and underlying influences. Analyzing relevance scores, saliency maps, and attention mechanisms offers retrospective insights. Yet, limited integration into the learning process confines their explanatory power, yielding partial insights into internal workings.

\begin{figure}
    \centering
    \includegraphics[width=\linewidth]{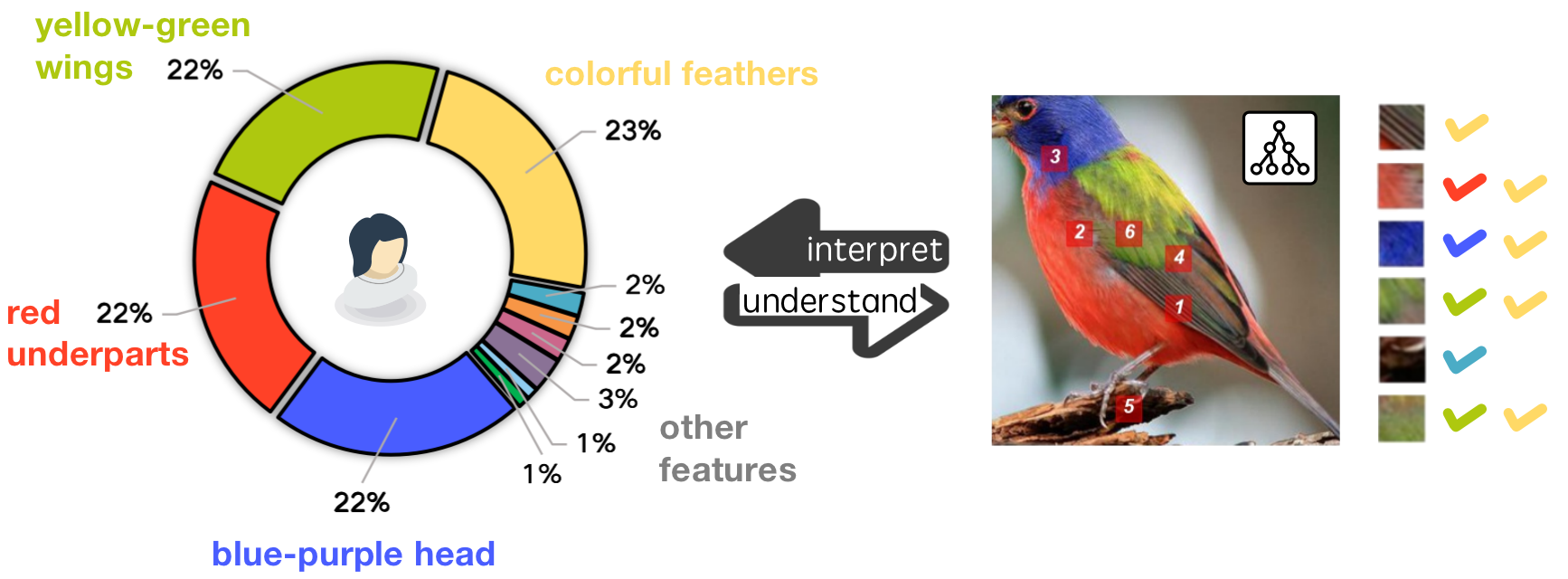}
    \caption{A comparative analysis of human and model focus in avian classification. Left: Visualization of human attention distribution during bird classification. Right: Decision path of \mymodell with highlighted patches, reflecting localized regions of interest. Conjunction of human focus and model patches are marked with checkmark, demonstrating consistency of cognitive approach between human and model.}
    \label{fig:enter-label}
\end{figure}

For direct model reasoning, decision trees integrate with neural networks, capitalizing on transparency. However, neural decision trees (NDTs) for fine-grained visual categorization (FGVC) have constraints~\cite{Li2022ASO}. One category, data-driven hierarchies, partitions data from coarse to fine, but induces training limitations. Moreover, prototypical approaches like ProtoTree~\cite{Nauta2020NeuralPT} and ViTNet~\cite{Kim2022ViTNeTIV} also face drawbacks. They lack step-wise reasoning, compromise interpretability with latent space projections, and forfeit interpretability by using soft trees with probabilities, nullifying single-path clarity.



In this study, we introduce ViTree, a single-path neural tree approach based on the vision transformer architecture, which utilize hard patches to achieve genuine step-wise representation learning while keeping state-of-the-art performance in fine-grained image classification tasks. 

Firstly, we introduce hard patches, serving as the human-understandable focuses of the model on critical local regions. These patches are selected at each decision tree node through a learnable process, representing locally significant features as perceived by the model. Subsequently, the model employs these patches to enhance image representation, constituting a round of representation learning when message passing between nodes. Importantly, these patches can be directly associated with the original image, enhancing intuitive and human-comprehensible inference compared to conventional soft methods reliant on prototypes.

Next we introduce the single path, or to say hard path. In this framework, each leaf node signifies a unique tree path. By generating diverse tree paths, ViTree facilitates comprehensive representation refinement within a broader feature space, thereby significantly enhancing its overall classification accuracy. Additionally, we introduce a meticulously designed leaf node selection module, enabling the model to effectively determine a single path among the tree's alternatives. This strategic selection process greatly facilitates the interpretability of ViTree, surpassing the capabilities of traditional ensembled forests or soft decision trees. 

Therefore, by following the chosen tree path, ViTree enables a transparent representation refinement process in a step-by-step manner with meaningful intermediate features. Consequently, ViTree serves as an avant-garde ante-hoc interpretable model.

Empirically, we substantiate the practical effectiveness of \mymodell through comprehensive experiments. Our extensive results corroborate that \mymodell attains state-of-the-art performance across diverse tiers of interpretable methodologies. Additionally, we employ multi-faceted approaches to corroborate the interpretability of our model, encompassing an algorithmic-level transparency analysis, an illustrative case study, and multiple surveys grounded in human-centered perspectives. These collective efforts, both theoretical and empirical, substantiate the interpretability of our approach.

Our main contributions are highlighted as following:
\begin{itemize}
    \item \textbf{Intrinsic Interpretation:} We infuse genuine representation learning into neural tree for FGVC. This grants each intermediate representation within tree nodes a profound meaning, enriching model interpretability.
    \item \textbf{``Hard"-Powered Simplicity:} Leveraging hard patches and paths during training and inference, we transcend the intricacies and ambiguity of previous SOFT methods, amplifying simplicity, clarity, and interpretability.
    \item \textbf{Unparalleled Performance:} Through rigorous experimentation, our model showcases state-of-the-art performance against formidable rivals.
    \item \textbf{Multi-perspective Interpretability:} Our model's interpretability is highlighted through diverse dimensions, especially pioneering human-in-the-loop surveys. These surveys vividly underscore \mymodel's innate human-understandability intuitively.
\end{itemize}

\section{Related Work}
\paragraph{Vision Transformers in FGVC.} 
Transformers~\cite{Vaswani2017AttentionIA, Dosovitskiy2020AnII} have showcased excellence in fine-grained visual classification. Notable examples include SIM-Trans~\cite{Sun2022SIMTransSI}, which integrates object structure, and ViT-SAC~\cite{Do2022FineGrainedVC}, addressing class uncertainty through self-assessment. IELT~\cite{Xu2023FineGrainedVC} unifies learning using ViT and ensemble techniques.
However, while attention maps provide a glimpse into the areas of focus within an image, they do not offer a direct and comprehensive understanding of the decision-making process, making it challenging for human observers to fully comprehend the underlying mechanisms driving the model's predictions. Addressing this interpretability challenge remains an ongoing research direction in the field.





\paragraph{Interpretability in FGVC.}
Prototype learning, led by ProtoPNet~\cite{NEURIPS2019_adf7ee2d}, stands as a prominent category of partial ante-hoc interpretable models for FGVC, because of their excellence at capturing subtle differences by generating prototypes that highlight specific local regions of an object. 
Various endeavors have aimed to enhance this approach, including prototype reduction~\cite{Rymarczyk2021InterpretableIC}, transformer adaptations~\cite{Xue2022ProtoPFormerCO}, boundary optimization~\cite{Wang2023LearningSA}, and deformable prototypes~\cite{Donnelly2021DeformablePA}.
However, a central drawback lies in potential loss of individual instance nuances when adopting prototypes. While these offer concise and interpretable data summaries, they might not fully encompass instance intricacies. This loss limits insights into the model's decision-making. Additionally, prototypes, constructed from an average of images in latent space, lack direct interpretable links to raw data. Consequently, prototypical methods offer limited human interpretability.

\paragraph{Neural Trees in Vision.}
Neural networks excel in capturing intricate relationships and robust representations, making them versatile tools in machine learning. Conversely, decision trees are valued for their sequential reasoning, enhancing model interpretability. In computer vision, neural trees harness these traits, yielding inherently interpretable models with strong performance. ANT~\cite{Tanno2018AdaptiveNT} integrates representation learning into adaptive architecture growth, while SeBoW~\cite{Chen2021SelfbornWF}, NBDT~\cite{Wan2021NBDTND}, and ACNet~\cite{Ji2019AttentionCB} derive tree structures from neural network decomposition, transformation, or incorporating convolution layers. For fine-grained visual categorization, ProtoTree~\cite{Nauta2020NeuralPT} and ViTNet~\cite{Kim2022ViTNeTIV} fuse prototype learning and neural decision trees under CNN and ViT backbones.
However, contemporary neural trees face challenges in reconciling performance and transparency~\cite{Akhtar2023ASO}, training robust hard trees, and ensuring interpretability of soft trees~\cite{Li2022ASO}. This study introduces \mymodel, a hard tree model with competitive performance, shedding light on these challenges.








\begin{figure*}[htbp]
    \centering
    \includegraphics[width=\linewidth]{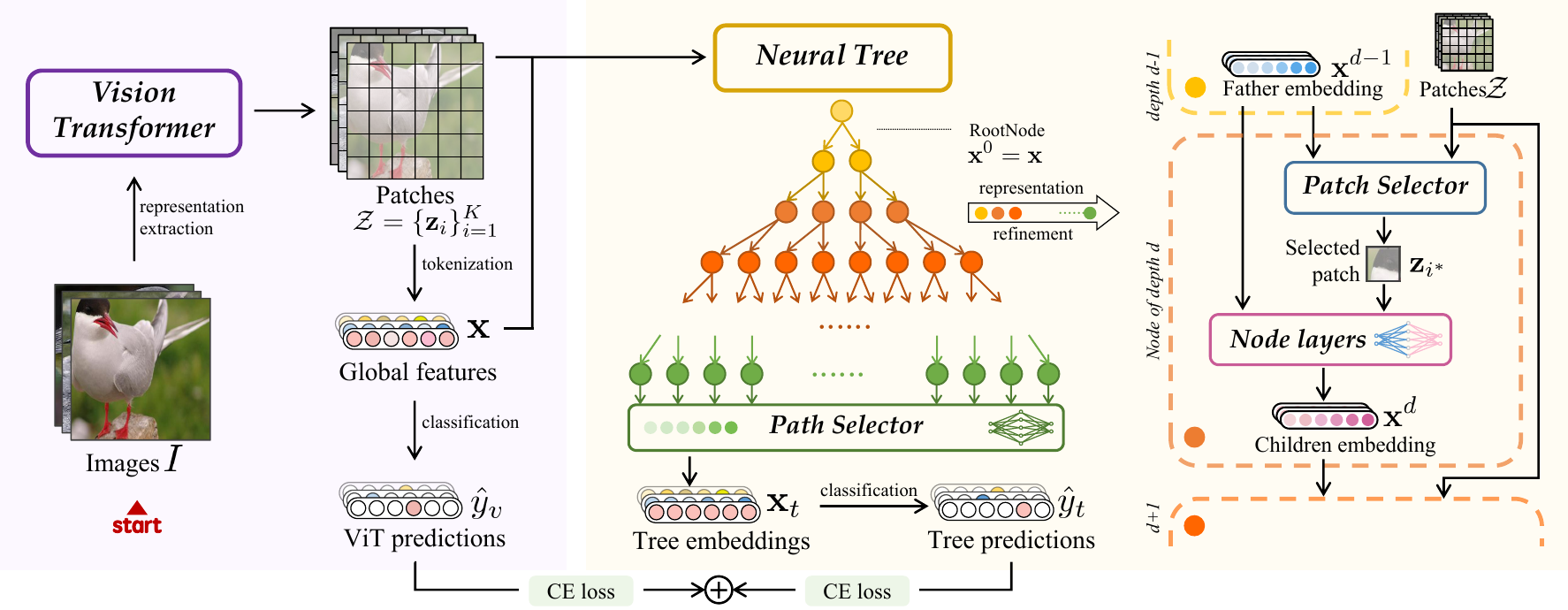}
    \caption{Illustration of the proposed \mymodell pipeline. Purple: The vision transformer module. It takes the raw images as input and output primary extracted features. Yellow: The neural tree module. The left part is a sketch of a tree and the right part is an example of parent-to-child representation learning process.}
    \label{fig:model}
\end{figure*}

\section{Methodology}
In this section, we begin by comprehensively analyzing \mymodel's structure and data flow. We then delve into the detailed design and implementation of each module in \mymodell. Finally, we explain the training methodology, providing a thorough understanding of the entire system.

\subsection{\mymodell Two-stage Framework}
As shown in Figure~\ref{fig:model}, a \mymodell model is composed of two parts: a vision transformer module and a decision tree module, where the former extracts informative but primary representation and the latter optimizes the representation along the tree paths. 

Let $I$ denote an image and $y$ is its ground-truth label. The backbone vision transformer \texttt{V} takes image $I$ as input, partitions it into a set of image patches, each of which corresponds to a specific region in the image, and obtains a set of patch-level representations $\mathcal{Z}=\{\mathbf{z}_i\}_{i=1}^K$ accordingly, as well as an image-level representation $\mathbf{x}$. By default, we use \textit{SwinT~\cite{Liu2021SwinTH}} as our vision transformer backbone, in which average pooling is used to agglomerate the patch-level representations into image-level representation. In some other transformers such as ViT~\cite{Dosovitskiy2020AnII}, a class token is set up in advance whose embedding is regarded as the image-level representation. Besides, a classification head is used to output a prediction $\hat{y}_v$, which will be used in later loss computation.

\begin{equation}
    \mathcal{Z}, \mathbf{x}, \hat{y}_v = \texttt{V}(I)
\end{equation}



Afterwards, $\mathcal{Z}$ and $\mathbf{x}$ are fed into the neural tree module $\texttt{T}$, and specifically, $\mathbf{x}$ serves as the embedding of the root node. Other than node respective embeddings, each node of $\texttt{T}$ contains a patch selector and several neural network layers, acting as a representation modifier. Through the process of parent-to-child information dissemination in the context of hierarchical tree structure, we optimize $\mathbf{x}$ in a representation learning way. Concretely, father nodes send their embeddings to children nodes. And next, children nodes select a patch from patches $\mathcal{Z}$ based on father embedding, then use the selected patch combining with father embedding to learn a optimized representation as children embedding. At last, the embeddings in the leaf nodes are sent into path selector module. The path selector picks one leaf and its node embedding is used as an output tree representation $\mathbf{x}_t$. Similarly, we get a tree prediction $\hat{y}_t$, which will also be used in loss computation.


\begin{equation}
    \mathbf{x}_t, \hat{y}_t = \texttt{T}(\mathcal{Z}, \mathbf{x})
\end{equation}

Generally speaking, the tree functions by constantly selecting local descriptive regions to provide guidance towards the progressive refinement of the primary features. Comparing with previous tree-based method~\cite{Nauta2020NeuralPT, Kim2022ViTNeTIV}, \mymodell has two main advantages in achieving human-understandable decision making process:
\begin{itemize}
    \item Hard patch: Unlike soft prototypes derived from extensive smoothing of training images, \mymodell adopts direct patches solely from the original image, enabling explicit observation of the model's operations.
    \item Hard path: Unlike utilizing soft distribution summation with path probabilities, \mymodell employs single path training and inference. Simultaneously, the model demonstrates sustained high and even better performance.
\end{itemize}

\subsection{Nodes and Representation Learning}
The nodes of the tree play a crucial role in representation learning. As shown in Figure~\ref{fig:model} (right), each node contains a representation of the current image. Different nodes within the same layer represent distinct directions of representation optimization. As information propagates deeper into the tree, the representation learning space expands, incorporating additional critical local information. Consequently, optimization leads to enhanced representations. Denote the node embedding of a father node at depth $d-1$ as $\mathbf{x}^{d-1}$, then we can derive the child embeddings $\mathbf{x}^{d}$ by:
\begin{equation}
\label{eq:patch}
    \mathbf{z}_i^\ast = \texttt{PatchSelector}(\mathbf{x}^{d-1},\mathcal{Z})
\end{equation}
\begin{equation}
    \mathbf{x}^{d} = \texttt{NodeLayers}(\mathbf{x}^{d-1},\mathbf{z}_i^\ast).
\end{equation}
The detailed composition of the modules can be found in later sections. Note that a father node can have several child nodes and the process of each child node is mutually independent to those of the others. Suppose the tree has $N_d$ leaves at depth $d$, then we choose one among them to output the final tree representation utilizing the path selector.
\begin{equation}
    \mathbf{x}_{t} = \texttt{PathSelector}(\{\mathbf{x}^{d,n}\}_{n=1}^{N_d}),
\end{equation}
where $\mathbf{x}^{d,n}$ is the node embedding of the $n$-th leaf depth $d$.

\subsubsection{Patch Selector.}
As described in Equation~\ref{eq:patch}, each tree node contains a patch selector that picks a single patch $\mathbf{z}_i^\ast$ from the image feature patches $\mathcal Z$, based on the input father embedding $\mathbf{x}^{d-1}$, for subsequent representation refinement. 

Firstly, we generated a patch weight map $\mathcal{W}=\{\mathbf{w}_i\}_{i=1}^K$ which assesses the importance of patches. The multi-head attention (MHA) mechanism as we used, or alternative methods, such as similarity-based measurements, weighted cosine, or linear learnable projections, could be considered for the weight generation process.

\begin{equation}
    \mathcal{W} = \texttt{MHA}(\mathbf{x}^{d-1}, \mathcal{Z}) \\
\end{equation}

Next, we pick the highest weighted patch $\mathbf{z}_{i^\ast}$ from patches $\mathcal Z$ based on the weight map $\mathcal{W}$ as described in the following equations. This patch $\mathbf{z}_{i^\ast}$ is then selected as the patch of the tree node, where
\begin{equation}
    i^\ast = \arg\max_{i} \{\mathbf{w}_i\}_{i=1}^K
\end{equation}

The patch selector aims to learn the most informative patch during message passing, corresponding to a significant region in the original image. Analyzing the patches chosen by the tree nodes provides valuable insights into the model's reasoning process in a human-interpretable manner.

\subsubsection{Path Selector.}
When the representations reach the lowest layer of the tree, each leaf node represents a unique learning path from the root node to that leaf. Consequently, the representations within each leaf node are the result of distinct optimization along different paths. The path selector is responsible for selecting a single path from them by choosing a specific leaf node. The node embedding of this selected leaf node will serve as the image representation $\mathbf{x}_t$ processed by the neural tree.

Similarly, we first learn a weight distribution $\mathcal{W}_l = \{ \mathbf{w}_{l,i}\}_{i=1}^{N_d} $ among $N_d$ leaf nodes according to their embeddings.

\begin{equation}
\label{eq:leafweight}
    \mathcal{W}_l=\texttt{FC}(\{\mathbf{x}^{d,n}\}_{n=1}^{N_d})
\end{equation}

Subsequently, we choose the leaf node with the highest weight as the selected node, $\mathbf{x}_t = \mathbf{x}^{d,i^\ast}$, where
\begin{equation}
\label{eq:leafmax}
    i^\ast = \arg\max_{i} \{\mathbf{w}_{l,i}\}_{i=1}^K
\end{equation}

Compared with previous approaches that typically rely on summation of leaf nodes' outputs to obtain distributions or combine multiple trees to achieve higher performance, our single-path method significantly enhances human interpretability.

\subsection{Training and Inference}
In our proposed method, consisting of the vision transformer module and the neural tree module, we aim to optimize both modules effectively. To achieve this, we employ cross-entropy loss individually for the vision transformer prediction, denoted as $\hat{y}_v$, and the tree prediction, denoted as $\hat{y_t}$. Specifically, $\ell_{vit}$ is utilized to guide the vision transformer in generating informative and robust representations, and $\ell_{tree}$ is applied to enhance the tree's representation refinement capabilities. To enable back-propagation, we combine these two cross-entropy losses as described in Equation~\ref{eq:loss}. When making prediction, we also leverage the two predictions in a compositional way with a learnable ratio.

\begin{align}
\label{eq:loss}
    &\ell = \ell_{vit} + \ell_{leaf} = \texttt{CEloss}(\hat{y}_v, y) + \texttt{CEloss}(\hat y_t, y) 
\end{align}

In this way, our approach ensures effective optimization and usage for both modules, contributing to improved performance in our model.



\begin{table*}[]
\resizebox{\textwidth}{!}{
    \begin{tabular}{|c|ll|cc|}
    \hline
    \textbf{Interpretability} &
      \multicolumn{1}{c}{\textbf{Method}} &
      \textbf{Backbone} &
      \textbf{CUB} & \textbf{Cars} \\ \hline
    Backbones 
                & SwinT-B~\cite{Liu2021SwinTH} & - & 91.2 & 94.5 \\ \hline
    \multirow{5}{*}{None or Post-hoc} 
                & TransFG~\cite{He2021TransFGAT} & ViT-B & 91.7 & - \\
                & FFVT~\cite{Wang2021FeatureFV}       & ViT-B      & 91.6 & - \\
                & AFTrans~\cite{Zhang2021AFL}   & ViT-B      & 91.5 & 95.0 \\
                & SIM-Trans~\cite{Sun2022SIMTransSI} & ViT-B      & 91.8 & - \\
                & IELT~\cite{Xu2023FineGrainedVC}  & ViT-B      & 91.8 & - \\ \hline
    \multirow{5}{*}{\begin{tabular}[c]{@{}c@{}}Latent \\ Prototypes\end{tabular}} 
     &ProtoPNet†~\cite{NEURIPS2019_adf7ee2d}  & ResNet-50 & 81.1 & - \\
     & ProtoPool~\cite{Rymarczyk2021InterpretableIC} & ResNet-152 & 81.5  & 88.9 \\
     & Def. ProtoPNet†~\cite{Donnelly2021DeformablePA} & ResNet-50 & 87.8  & - \\
     & ST-ProtoPNet~\cite{Wang2023LearningSA}   & ResNet-50  & 88.0 & -  \\
     & ProtoPFormer~\cite{Xue2022ProtoPFormerCO}   & DeiT-Ti    & 82.3 & 88.4 \\ \hline
    Interpretable ML &
      SLDD-Model~\cite{Norrenbrock2023Take5I} & ResNet-50 & 86.5 & 93.3\\ \hline
     Neural Tree & ACNet~\cite{Ji2019AttentionCB} & ResNet-50  & 88.1 & 94.6 \\ \hline
    \multirow{2}{*}{Neural Tree + Prototypes} &
        ProtoTree†~\cite{Nauta2020NeuralPT} & ResNet-50 & 82.2 & 86.6\\
        & ViT-NeT~\cite{Kim2022ViTNeTIV}    & SwinT-B   & 91.6 & 95.0\\ \hline
    \multirow{2}{*}{Neural Tree + Hard Patches}
    & \multirow{2}{*}{ViTree(Ours)}
    & \multirow{2}{*}{SwinT-B}
        & \textbf{92.0} & \textbf{95.1} \\
        & & & (+0.8) & (+0.6)\\ \hline
    \end{tabular}
}
\caption{Top-1(\%) accuracy comparison on CUB-200-2011 and Standford Cars. \mymodell attains state-of-the-art performance on both datasets (marked in bold), enhancing the backbone's performance by 0.8\% and 0.5\% respectively (highlighted in red). The symbol † denotes that the reported result is derived from an ensemble of corresponding models (ProtoPNet or ProtoTree).}
\label{tab:result}
\end{table*}

\section{Experiments}
We evaluate \mymodell on several commonly used fine-grained image classificaton datasets. In this section, the experiment setup and quantitative results will be introduced.

\subsection{Experiment Setup}
\subsubsection{Datasets, Baselines.} We evaluated \mymodell on two benchmark datasets: CUB-200-2011~\cite{Wah2011TheCB} and Standford Cars~\cite{Krause2013CollectingAL}. 
We conducted a comparative analysis between \mymodell and several state-of-the-art methods, encompassing increasing levels of interpretability. As concluded in Table~\ref{tab:result}, these methods included transformer-only methods with SOTA performance and little interpretability, prototype-learning approaches, interpretable machine learning methods, neural tree methods, as well as combinations of neural tree and latent prototypes methods.


\subsubsection{Training Details.} 
\label{sec:detail}
For the vision transformer module, we mainly follow~\cite{Liu2021SwinTH} for SwinT implementation, making a little adaptation at the forward stage to output patches features before flattening. For fair comparison, we adopt the image preprocessing and backbone architecture methodologies outlined in~\cite{Kim2022ViTNeTIV}. For the decision tree module, we implement a binary tree with a depth $d=6$. We use multi-head attention in $\ \texttt{PatchSelector}$, and $\ \texttt{NodeLayers}\ $ module is composed of several fc layers, a RELU layer, a dropout layer and a batchnorm layer.
For all methods, we determine the hyperparameters, including learning
rate, weight decay, dropout, number of layers via heuristic experience and gird search.

\subsection{Results on Benchmarks}
\label{sec:result}
\begin{figure}[H]
        \centering
        \includegraphics[width=\linewidth]{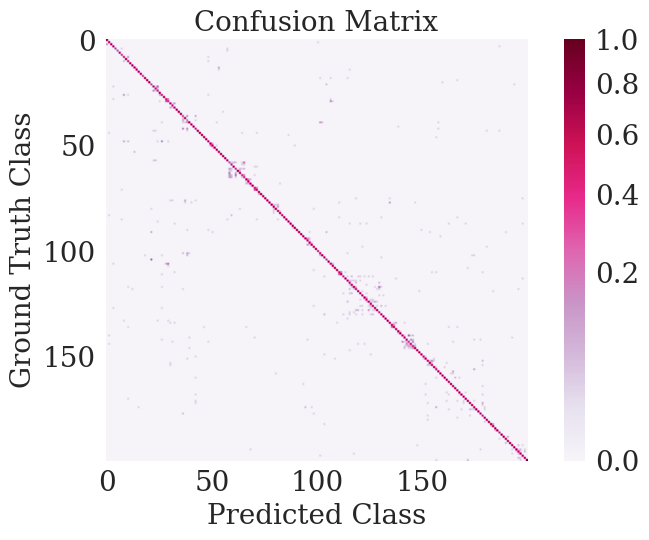}
        \caption{Confusion matrix on CUB-200-2011.}
        \label{fig:confusion}
    \end{figure}

Table \ref{tab:result} presents a comparison of the Top-1 accuracy achieved by \mymodell with various strong competitors. On the CUB-200-2011 dataset \cite{Wah2011TheCB}, \mymodell achieves a remarkable accuracy of 92.0\%, surpassing various levels of interpretable methods by a margin of 0.4\% to 10.9\%. High interpretability and high performance usually cannot coexist. Specifically, the transformer-only methods have little interpretability while displaying commonly higher performance. Despite the typical trade-off, our \mymodell consistently outperforms leading transformer-only methods by 0.2\% to 0.5\%, all while maintaining exceptional model transparency. This achievement underscores our model's capability of delicately balancing between interpretability and performance.
On the Stanford Cars dataset \cite{Krause2013CollectingAL}, \mymodell attains an accuracy of 95.1\%, outperforming state-of-the-art methods as well. Furthermore, we present the confusion matrix for our CUB-200-2011 results in Figure~\ref{fig:confusion}, where the pronounced predictive accuracy can be readily discerned. Our findings conclusively demonstrate that \mymodell achieves the highest performance while concurrently offering superior interpretability in fine-grained visual categorization. 

\begin{multicols}{2}[]
    
    \columnbreak
    
    \begin{figure}[H]
        \centering
        \includegraphics[width=0.85\linewidth]{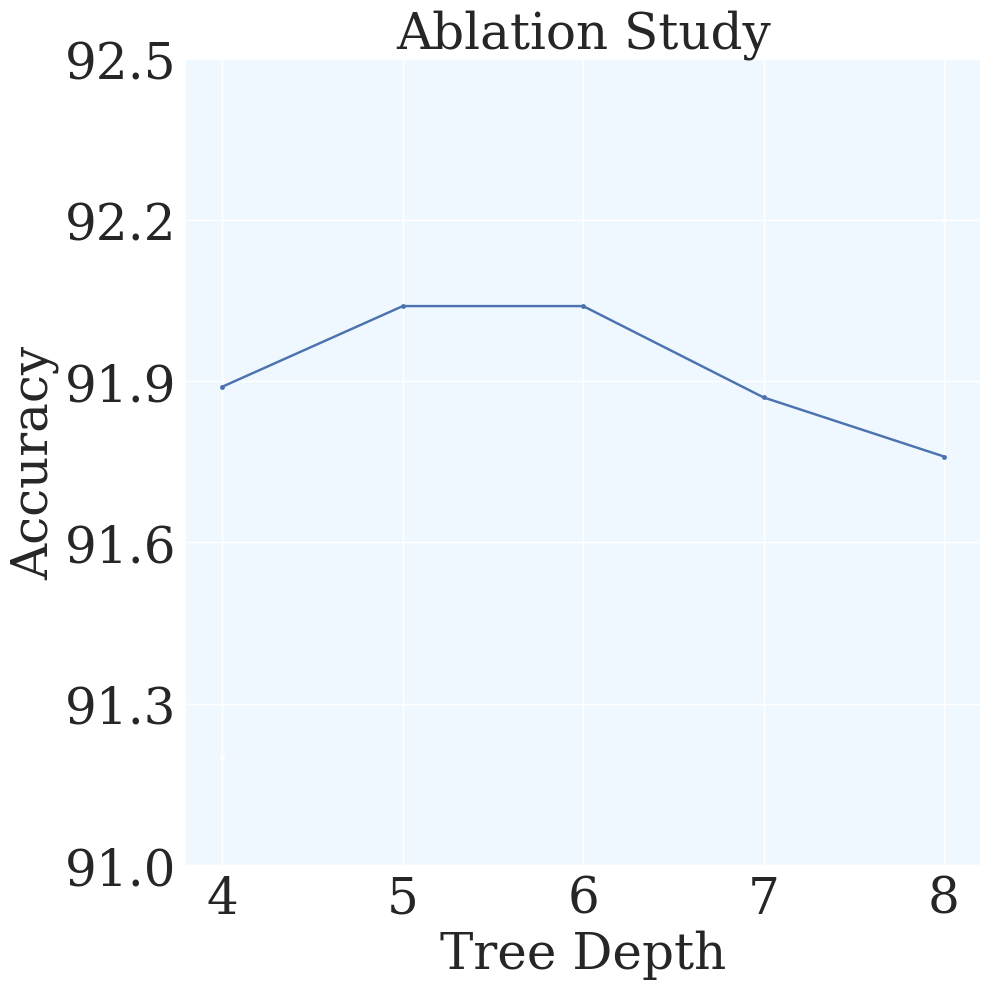}
        \caption{Effect of tree depth on CUB-200-2011.}
        \label{fig:treedepth}
    \end{figure}
    
\end{multicols}

\subsection{Ablation Study}
\subsubsection{Tree Depth.}
The architectural configuration of the tree significantly impacts the performance of the model. Consequently, we delve into the impact of tree depth on the CUB-200-2011 dataset. As depicted in Figure~\ref{fig:treedepth}, optimal performance manifests with a 5-layer or 6-layer neural tree. This observation can be attributed to the phenomenon whereby shallow trees lack the requisite predictive capacity to effectively explore the available learning space. Conversely, excessively deep trees prove challenging to train, susceptible to overfitting, and more sensitive to noise.

\subsubsection{Loss Component.} As depicted in Equation~\ref{eq:loss}, we employ a combinatorial training loss to cohesively optimize both the vision transformer and the neural tree module in our benchmark setting. To substantiate the efficacy of our approach, we assess the individual impact of each loss component, as showcased in Table~\ref{tab:ablation}, presenting the outcomes and comparisons. When utilizing $\ell = \ell_{\text{vit}}$, we report the accuracy of $\hat{y}_v$; Similarly for $\ell = \ell_{leaf}$, we present the accuracy of $\hat{y}_t$. The results underscore that the single-loss performance does not match the benchmark setting's performance. The integration of a combinatorial loss mechanism along with a compositional prediction strategy significantly augments the model's predictive capacity. This enhancement can be attributed to the synergetic effect of combining diverse loss components, enabling the model to capture intricate relationships within the data. The compositional prediction further refines the model's understanding, enabling it to harness the strengths of both the vision transformer and the neural tree module, ultimately leading to more accurate predictions.


\subsubsection{Leaf Strategy.}
In our implementation, we leverage a learned weight (Equation~\ref{eq:leafweight}) to evaluate the significance of the leaves and choose the highest weighted leaf (Equation~\ref{eq:leafmax}). We term this approach succinctly as ``Learn - Hard," signifying the selection of a single robust decision path through learnable weights. Beyond the interpretability benefits, this ablation study aims to empirically establish the superiority of our strategy in enhancing model performance. 

As depicted in Table~\ref{tab:ablation}, we present the top-1 accuracy results across various leaf strategies. The ``Mean" strategy entails the averaging of all leaf representations for the output. ``Prob." denotes the utilization of path probabilities, wherein we compute a distribution of probabilities among leaves based on patch similarities along corresponding tree paths. This probability serves as the weight for each leaf, obviating the need for explicit learning. This is a widely adopted mechanism aimed at bolstering model performance previously~\cite{Wan2021NBDTND}. Lastly, ``Hard" and ``Soft" signify the choice between selecting the leaf with the maximum weight and aggregating all leaves with respective weights. 

The results indicate that the ``Hard" solutions surpass their corresponding ``Soft" counterparts, as well as the ``Mean" strategy. This observation underscores our model's exceptional representational prowess, as a single path proves sufficient for accurate predictions. Furthermore, our observations reveal that the ``Learn" strategies outperform the ``Probability" strategies. This distinction may be attributed to the inherent flexibility imparted by the learning weight strategy, endowing the model with enhanced potential.

\begin{table}
\resizebox{0.48\textwidth}{!}{
    \begin{tabular}{|cl|cc|}
\hline
\multicolumn{2}{|c|}{\textbf{Ablation settings}} & \multicolumn{1}{c}{\textbf{Top-1(\%) Acc.}} & \multicolumn{1}{c|}{\textbf{Degradation}} \\ \hline
\multicolumn{1}{|c}{\multirow{2}{*}{Loss}}                               
& $\ell = \ell_{vit}$ &   91.66   &   -0.37     \\
& $\ell = \ell_{leaf}$ &    91.61      &  -0.42   \\ \hline
\multirow{5}{*}{\begin{tabular}[c]{@{}c@{}}Leaf\\ Strategy\end{tabular}} 
& Mean  &   91.57    &   -0.46   \\                 
& Prob. - Soft &     91.52    &  -0.51     \\
& Prob. - Hard  &     91.73    &  -0.30   \\
& Learn - Soft &     91.40     &  -0.63     \\
& Learn - Hard &      92.03     &    /   \\ \hline
\end{tabular}
}
\caption{Ablation study on loss and leaf strategy regarding the CUB-200-2011 dataset. Results are compared with those acquired from the benchmark experimental setting.}
\label{tab:ablation}
\end{table}

\begin{table*}\small
\centering
    \begin{tabular}{|l|c|c|c|c|}
    \hline
    & ACNet & ProtoTree & ViTNeT & ViTree \\ \hline
    Year & 2020  & 2021      & 2022   & 2023   \\ \hline
    Single path inference during training and testing & - & - & - & + \\ \hline
    Non-ensemble tree instead of usage of forest & + & + & + & +\\ \hline
    Genuine learner performing step-wise representation learning while making decisions  & - & - & - & +  \\ \hline
    Bigot leaves where each leaf represents predetermined classes distribution& - & + & - & - \\ \hline
    Transparent router with hand-crafted mechanism instead of black-box neural network & - &   + &  + & + \\ \hline
    Other contributions such as reasoning with smoothed prototypes or more direct hard patches & - &   + &  + &  + +\\ \hline
    \end{tabular}
\caption{Qualitative comparison between recent works of different neural trees, with respect to their interpretability.}
\label{tab:compare}
\end{table*}

\section{Interpretability and Visualization}

In this section, we undertake a comprehensive assessment of the interpretability of \mymodell through empirical comparisons, case study and human-centered survey. We always utilize the model trained on CUB-200-2011 dataset as a representative example.

\subsection{Empirical Comparison}
In accordance with~\cite{Li2022ASO}, we assess the interpretability of \mymodel, alongside several prominent previous approaches using several qualitative indicators as in Table~\ref{tab:compare}. 


Note that \mymodell offers remarkable advantages in terms of model interpretability. One of its unique capabilities is achieving single-path training and inference. Unlike conventional approaches that rely on distributions of predictions or path probabilities to enhance model performance, \mymodell outperforms them by using a non-compositional decision path. This demonstrates its excellent feature extraction and key region recognition capabilities.

Another distinguishing feature of \mymodell is its support for step-wise representation learning. Through the integration of patches along the tree path, \mymodell achieves genuine representation learning rather than relying on mere transformations with networks. Taking each tree path, it can provide meaningful intermediate results at any tree layers, representing different extent of representation refinement.

Furthermore, \mymodell employs hard patches as an auxiliary support, contributing to its superiority in human-understandability compared to using prototypes. Hard patches are more easily understandable since they come from the same image as the features to be refined, whereas prototypes are derived from multiple images in the training set and represent abstract characteristics.


\subsection{Case Study}
\begin{figure*}
    \centering
    \includegraphics[width=0.95\linewidth]{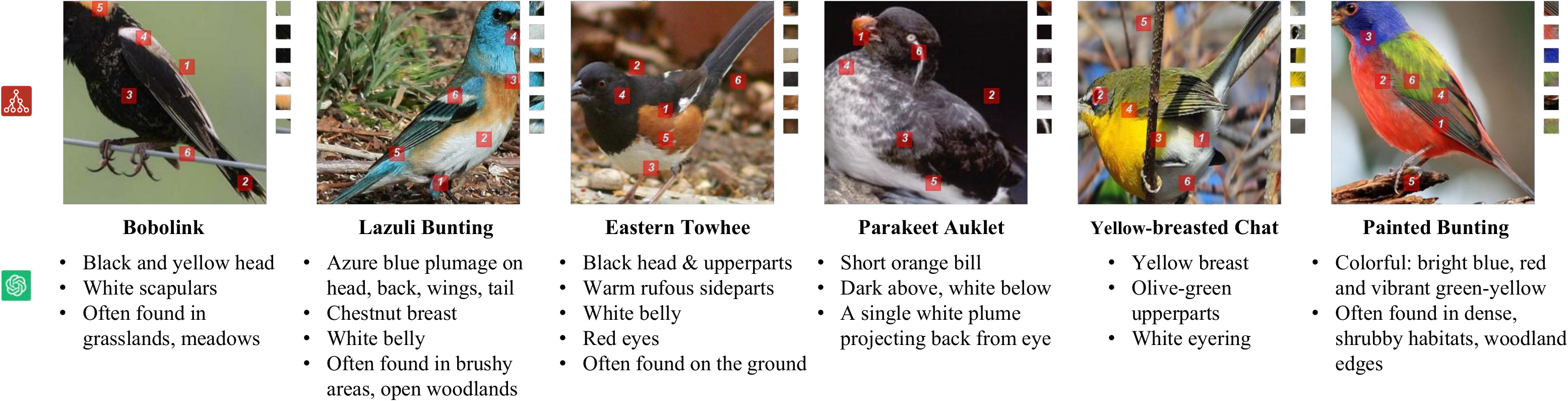}
    \caption{ViTree's proficiency in capturing key classifying attributes in accordance with human among bird species: The upper part illustrates the model's patch selections along the decision path. Highlighted in red, these selected patches are labeled with their order on the image, and listed to the right of the figure. The lower part provides a summary of ChatGPT's insights on distinctive human-observable traits for each species. Notably, these traits align closely with our model's focal points, underscoring a robust harmony between our model's internal logic and human perspectives.}
    \label{fig:casestudy}
\end{figure*}

Figure~\ref{fig:casestudy} illustrates our model's decision-making process using select examples, offering an intuitive depiction of its operation. To corroborate our model's alignment with human perceptual judgments, we engage ChatGPT~\cite{openai2023chatgpt} to succinctly summarize salient distinguishing attributes.

Evidently, a significant portion of our selected patches converge upon avian subjects, aptly capturing their essential traits like colors and patterns. Occasional patch selections encompass the background, signifying the informative role of environmental context in the model's decisions. This phenomenon likely emanates from the diverse ecological habitats that various avian species inhabit, contributing as discriminative features. For instance, as articulated by ChatGPT, the bobolink thrives in grasslands, the eastern towhee prefers terrestrial environments, while the painted bunting finds refuge in woodlands. This spatial affinity aligns with patches occasionally focusing on background details in their respective decision paths.


\subsection{Human-centered Survey}
Consistent with~\cite{Wan2021NBDTND}, we adhere to the approach outlined by~\cite{PoursabziSangdeh2018ManipulatingAM} to establish a benchmark for quantifying model interpretability from a human-centric standpoint. This benchmark asserts that individuals should possess the ability to replicate and willingly trust a model's predictions, while also being capable of identifying any errors made. Following above criterion to substantiate the interpretability of \mymodel, we have formulated two distinct surveys, detailed as follows.

\subsubsection{Explanation-guided Level of Trust.}
In this comprehensive survey, participants are presented with randomly selected images from the training set to establish a preliminary understanding of avian features, enabling them to develop individual focal points for species identification. Subsequently, a test image, accompanied by the model's decision path, is provided, prompting participants to assess the model's judgment process. Remarkably, the model achieves accurate predictions across all inquiries.

The findings reveal that 470 out of 671 (70.04\%) responses acknowledge comprehending the decision path, demonstrating the model's interpretability. Participants identify a shared analytical pattern between humans and the model: a holistic assessment followed by localized feature recognition. Both parties converge on intuitive and salient traits as focal points. However, individuals unable to grasp the model's decision-making process attribute this discrepancy to the model's inclination to consider background information, a characteristic distinct from human focus. This phenomenon is explained by the model's capability to discern habitat-related cues.


\subsubsection{Identifying \mymodell Mistakes.}

In this conducted survey, we adhere to the methodology outlined by Wan et al.~\cite{Wan2021NBDTND}. Participants are tasked with identifying incorrect predictions from a randomized combination of scenarios encompassing two instances of correct predictions and one instance of an erroneous prediction. The images, coupled with the model's decision path but without the ultimate predictions, are presented for assessment. The results demonstrate that 421 out of 513 (82.06\%) responses accurately detect the model's error.

Subsequently, through an analysis of the model's mistakes, we ascertain their comprehensibility. To further investigate, a subsequent survey is devised. This survey involves the presentation of misclassified images alongside an image from the corresponding misclassified class, identified using the Faiss library~\cite{faiss}. The image pairs are presented without the model's decision path. Participants are tasked with determining whether the paired images belong to the same or different classes. Control groups are established, wherein image pairs are sampled randomly from the same class. The outcomes reveal that for pairs comprising misclassified and mistaken-class images, 529 out of 726 (72.87\%) responses deem the pairs to belong to the same class, while 197 out of 726 (27.13\%) discern the discrepant classification. Conversely, for control pairs, 66 out of 264 (25.00\%) responses perceive incongruence in class assignment, while 198 out of 264 (75.00\%) accurately identify the shared classification. Notably, the ratios within the two aforementioned groups display consistent patterns.

The survey outcomes distinctly establish that, given the model's decision path, participants adeptly discern model errors. Conversely, when the model's rationale is absent, individuals tend to replicate model inaccuracies. These findings substantiate two pivotal observations. Firstly, participants' ability to comprehend model errors strongly correlates with the provided decision path, thus highlighting their dependency and trust on the decision paths, and also aptitude in evaluating the correctness of decision paths. Secondly, the model's error patterns closely resemble human error tendencies, underscoring the achievement in emulating human cognition to inform model design.

\section{Conclusion}
In this paper, we proposed \mymodel, a single-path neural tree combined with vision transformer method which achieves interpretable fine-grained visual categorization with step-wise representation learning. The utilization of hard patches and selection of a single hard tree leaf strengthen model's inner transparency. Meanwhile, extensive experiments validate the state-of-the-art performance of \mymodel, and multi-perspective methods proves its human understandability. In the future, we anticipate a trajectory of continuous model refinement, where interpretability and performance harmoniously advance, and we wish for somedays, machine intelligence can be seamlessly intertwined with human insight, ultimately enriching both realms.

\section{Acknowledgements}
This work was supported by NSFC 62176159, 62322604, Natural Science Foundation of Shanghai 21ZR1432200, and Shanghai Municipal Science and Technology Major Project 2021SHZDZX0102.

\newpage
\bibliography{AnonymousSubmission/LaTeX/aaai24}



\end{document}